\documentclass[preprint]{elsarticle}

\usepackage{hyperref}
\usepackage{algorithm,algpseudocode}
\usepackage{amssymb}

\journal{Journal of \LaTeX\ Templates}

\begin{document}

\begin{frontmatter}

\title{Solving Area Coverage Problem with UAVs: A Vehicle Routing with Time Windows Variation}

\author[mymainaddress]{Fatih Semiz\corref{mycorrespondingauthor}}
\cortext[mycorrespondingauthor]{Corresponding author}
\ead{fsemiz@ceng.metu.edu.tr}

\author[mymainaddress]{Faruk Polat}
\address[mymainaddress]{Computer Engineering Department, Middle East Technical University, 06800, Cankaya, Ankara, Turkey}

\begin{abstract}
In real life, providing security for a set of large areas by covering the area with Unmanned Aerial Vehicles (UAVs) is a difficult problem that consist of  multiple objectives. These difficulties are even greater if the area coverage must continue throughout a specific time window. We address this by considering a Vehicle Routing Problem with Time Windows (VRPTW) variation in which capacity of agents is one and each customer (target area) must be supplied with more than one vehicles simultaneously without violating time windows. In this problem, our aim is to find a way to cover all areas with the necessary number of UAVs during the time windows, minimize the total distance traveled, and provide a fast solution by satisfying the additional constraint that each agent has limited fuel. We present a novel algorithm that relies on clustering the target areas according to their time windows, and then incrementally generating transportation problems with each cluster and the ready UAVs. Then we solve transportation problems with the simplex algorithm to generate the solution. The performance of the proposed algorithm and other implemented algorithms to compare the solution quality is evaluated on example scenarios with practical problem sizes.
\end{abstract}


\end{frontmatter}

\section{Introduction}
The vehicle routing problem (VRP) is a famous combinatorial optimization problem proposed in the late 1950s. In VRP aim is to deliver goods or services to $n$ customers by $m$ vehicles that are initially located at a depot. While achieving this, the optimal set of routes need to be found for the fleet. The objective is to minimize the total operation cost as well as satisfying the constraint that each customer is served only once. There is still a great interest in the VRP because it is both practically important and it is difficult to solve. Finding an optimal solution to VRP is shown to be NP-Hard ~\cite{VRP-NpHard}, hence the size of the problems that can be solved optimally may be limited. There are many studies on solving VRPs optimally or near-optimally for large scale problem instances. However, for some cases, classical VRP formulation is inadequate to represent real-world scenarios. Because of that VRP has several variations, Faied et al. \cite{VRPvariations} provided a survey about these variations. One variation of VRP is capacitated vehicle routing problem (CVRP). In this setting, the vehicles have limited carrying capacity of the goods. Another variation is the vehicle routing problem with time windows (VRPTW). In this setting, the service time of each delivery location is limited within a time window $[a_{i}, b_{i}]$. 

\begin{figure}[!h]       
  \begin{center}    
 \includegraphics[scale=0.6]{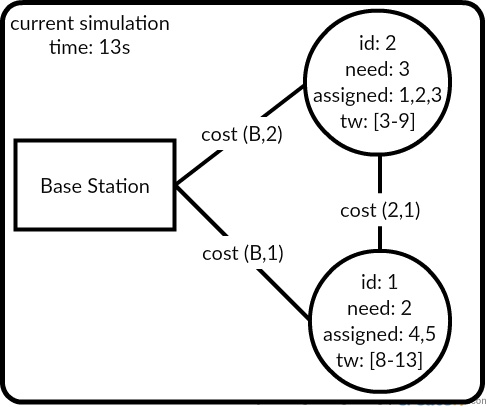}    
    \caption{BURAYA PROBLEMIN BASLANGICINI GOSTEREN YENI CIZIM GELICEK}\label{Figure::Ornek1}        
  \end{center}     
\end{figure}

Motivated by these variations we also provided a variation which solves an important real-world problem. In this study, we aimed to provide security for a set of large areas by UAVs. In the problem set, we have a fleet of UAVs to complete the mission. In the beginning, all of them are placed in a depot. Our task is to monitor a set of terrains. Each terrain has a different size. Hence, we have to cover each terrain with a specific number of UAVs. Moreover, each terrain has a specific time window to cover. The UAVs must remain at the target areas from the beginning of its time window till the end of its time window and UAVs may visit other target areas after they finish their jobs. Other modern examples of this problem include aerial mapping, oil and gas industry, broadcasting, disaster managing, and homeland security ~\cite{Bigs,doi:10.7901/2169-3358-2008-1-113,surveyDisaster,surveyDisaster2,military}. A representation of a MAPF-MD problem is provided in Figure-\ref{Figure::Ornek1}.

In this Figure, there are 6 UAVs positioned at a depot and there are 3 areas to be covered. The quadcopters represent the UAVs and the houses represent the areas to be covered. The numbers inside the houses represent the number of UAVs needed to cover that area and the numbers behind the houses represent the time windows of those targets. The aim is to cover all 3 areas by assigning enough number of UAVs. The UAVs must remain at the areas until the time window of that target ends.

We represented this problem as a VRPTW application and solved it with a novel approach. The main idea is to cluster the target areas according to their time windows. Then the clustered targets and the UAVs ready in that time are converted into transportation problem~\cite{TransportationProblemBook} and the transportation problem is solved with the simplex algorithm \cite{SimplexBook}. After solving each cluster we generate the new transportation problem with the ready UAVs and the next cluster and continue solving them with the simplex algorithm. In the end, we aggregated the solutions to generate the overall solution. 

We implemented the suggested method and four other alternative methods to compare the performance of our algorithm. We tested our solution in two different data sets the first one consists of the handcrafted scenario and the second one consists of randomly created scenario. For both of them, we used $20\times20$ grid environment 5 identical UAVs and varying size of targets (3, 5, 7, 10, 25 target scenarios). In randomly created scenarios we randomly determined the time windows and the coordinates of the agents. In both sets, we performed 100 different experiments. Our contributions to this field were the VRPTW variation for the UAV area coverage problem and the novel algorithm defined to solve this problem that returns fast, good quality solutions.
\section{Related Work}
The problems that are related to our problem can be classified into three categories: the VRP problem, the VRPTW problem and its variations, and VRP instances with applications to UAV operations.

VRP problem first appeared at Dantzig and Ramser's study in 1959 \cite{doi:10.1287/mnsc.6.1.80}. In this study they tried to find the optimum routing for a fleet of gasoline trucks that are making petrol transfer. They suggested an integer linear programming approach that finds near optimal solutions. In 1992 Laporte \cite{LAPORTE1992345} he provided a survey about exact and approximate algorithms developed for the VRP. In that study he provided 6 different examples of exact algorithms and 4 different examples of heuristics. A more recent survey is performed by Kumar and Panneerselvam \cite{VrpSurveyKumar} in 2012. In this study they stated that some examples of the exact algorithms to solve VRP problem include; the branch-and-bound algorithm, the branch-and-cut algorithm, and the branch-and-price algorithms. For the heuristic algorithms they mentioned cluster-first route-second algorithms, route-first cluster second algorithms and some metaheuristics like constraint programming, simulated annealing and tabu search. In the following paragraphs we will first  define some common VRP variations and then provide some examples of exact and heuristic algorithms.

Some common VRP variations are \cite{VRPvariations}:
\begin{enumerate}
\item \textit{Capacitated VRP (CVRP):} In this problem each vehicle has a uniform capacity $Q > 0$ and each customer has a capacity $q_{i}$ such as $Q \geq q_{i} > 0$. Each vehicle must be served by a single vehicle and no vehicles can serve a set of customers who exceeds its capacity \cite{Lysgaard2004}.
\item \textit{Vehicle Routing Problem with Time Windows (VRPTW):} In this VRP variation each delivery location has a specific time window assigned to it and they must be visited inside these time windows \cite{AZI2010756}.
\item \textit{The Vehicle Routing Problem with Pick-up and Delivery (VRPPD):} In this VRP variation clients require simultaneous pick-up and delivery service \cite{ALFREDOTANGMONTANE2006595}.
\item \textit{Split Delivery VRP (SDVRP):} In this VRP variation each customer can be served by different vehicles if it reduces operation cost \cite{doi:10.1287/trsc.1040.0103}.
\end{enumerate}

Our problem is a variation of VRPTW problem in which agents must remain at the target location until its time window ends. Also, the targets must be visited by more than one agent which is similar to CVRP, and targets may be visited by different vehicles which is close to SDVRP problem. Because of these features, we mainly analyzed approaches solving VRPTW problem variations, and provided some examples of CVRP and SDVRP solutions.

Recent studies generally used the branch and bound algorithms to find exact solutions to the problems with large state spaces~\cite{articleClausen}. Branch and bound algorithm (B$\&$B) represent the unexplored subspaces as nodes of a dynamically generated search tree. At each iteration branch and bound algorithm creates one such node. It uses a bound function to calculate the expected value of best possible node from a sub-tree. If the bound is worse than the current state, it discards that sub-tree. Otherwise, B$\&$B adds that sub-tree to the pool of live nodes with its bound information. This process continues until one node is left which is the optimal node. In 1997, Kohl and Madsen~\cite{725f9b56d0e64197a2b79585c9409f49} solved VRPTW optimally using the branch and bound algorithm. Guti\'errez-Jarpa et al.~\cite{journals/eor/Gutierrez-JarpaDLM10} optimally solves VRPTW by a branch and price algorithm. In this algorithm, at each node of the search tree, columns may be added to the linear programming relaxation. Another branch in the exact algorithm family is the methods using set-partitioning (SP) formulation. This formulation is originally published by~\cite{Balinski:1964:IPD:2783665.2783675}. In this study they solved CVRP by representing a route by a binary variable. The aim is to minimize that binary variable to find the optimal path. Contardo and Martinelli~\cite{CONTARDO2014129}, solved multi-depot vehicle routing problem (MDVRP) by using a vehicle-flow and a set-partitioning formulation. They exploited both of the techniques at different stages of the algorithm. In the study~\cite{Mateus-set-part}, authors provided a two-level algorithm using the genetic algorithm and SP together to solve VRPTW. 

In 2006 Chen et al.~\cite{CHEN2006383} solved real-time time-dependent vehicle routing problem with time windows. They formulated the problem as a series of mixed integer linear programming instances. Variables in the formulation are vehicle routes and departure times. Authors proposed route-construction and route improvement heuristics to solve the problem. Lau et al. solved VRPTW and m-VRPTW (VRPTW with limited number of vehicles) with a tabu search in~\cite{LAU2003559} which is a metaheuristic. A metaheuristic is a higher-level procedure which is used to select a partial search algorithm that may provide sufficiently good solutions for an optimization problem. Authors calculated an upper bound for this problem setting and showed that their solution is close to that upper bound. Montemanni et al. solved dynamical vehicle routing problem (DVRP) with an ant colony system (ACS)~\cite{Montemanni2005}. In DVRP new tasks arrive dynamically and new tasks can be assigned to the vehicles that already left the depot. The main idea of their study is slicing the day into time periods. They used these time periods to generate a sequence of static vehicle routing problems. They solved each of these static VRPs with ACS. Whenever they found a good solution in a time period, they transferred that information to the next time period. They defined a public domain which includes new benchmark problem instances. They showed that their solution yields good solutions on both artificial and real problems. Another interesting approach using meta-heuristics is ~\cite{Alba:2005:ETD:2221377.2221717}. In this study, a population-based heuristic is used for solving CVRP. As a heuristic, they proposed a cellular Genetic Algorithm (cGA). They provided high-quality solutions with this study and improved several existing results. In 2014, Ray et al.~\cite{RAY2014238} solved the multi-depot split-delivery vehicle routing problem (MDSDVRP) with a new integer linear programming (ILP) model. They provided a fast heuristic algorithm enriching knowledge gathering to find near-optimal solutions.

UAV path planning problem has several variations. These variations include different type of time domains and environmental models. A method can solve UAV path planning in  online fashion or offline fashion. Moreover, environmental models of the problem can also vary. There are 2D and 3D variations of UAV path planning problem. Zhao et al.~\cite{ZHAO201854} provided a survey of studies solving different type of UAV planning problems.
\section{Problem Description}
VRPTW problem instance studied in this project consists of $m$ UAVs where each of them is identical, has a capacity of one and has a predefined fuel constraint. The UAV list is represented with $A = (a_{1}, a_{2} \cdots a_{m})$ and the UAVs are represented with $a_{i}$. The problem also consists an undirected graph $G=(V, E)$. $V = \{v_{1},v_{2},v_{3} \cdots v_{n}\}$ is the set of target locations where $v_{1}$ is the depot, and $E = \{(e_{i}, e_{j}),i,j \in V,i \neq j\}$ is the edge set. Each target $v_{i}$ must be visited within its time window and the needed number of vehicles (number of UAVs needed to cover the area) must remain at that target throughout the time window. A typical time window for $i_{th}$ target is defined by $T_{i} = (t_{s},t_{f})$ where $t_{s}$ represents start of the visiting time and $t_{f}$ represents end of visiting time. UAVs staying at a target for a time window assumed to spend a constant amount of fuel. An UAV with enough fuel can be assigned to another target after the time window of the previously visited target finishes. $C$ represents the set of costs between locations where $c_{ij}$ is the travel cost between locations $i$ and $j$. The UAVs must start their job from the depot ($v_{1}$) and end at the depot. The goal is to find a scheduling strategy that visits each location with the necessary number of UAVs throughout time windows while minimizing the total distance traveled. The solution must satisfy the time window and fuel constraints.

\begin{figure}[htpb]       
  \begin{center}    
 \includegraphics[scale=0.4]{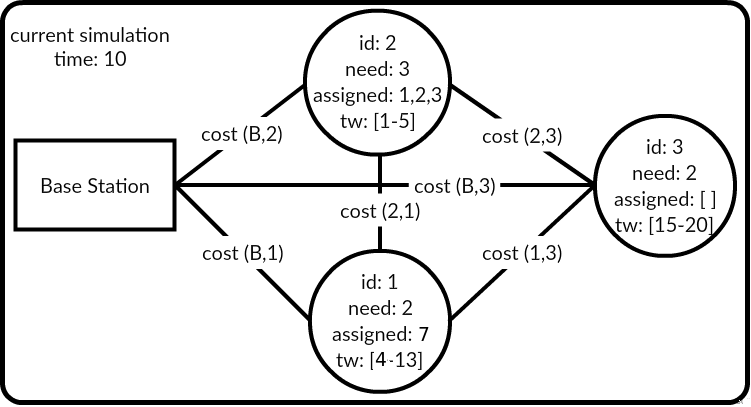}    
    \caption{A sample drawing of the studied problem}\label{Figure::ProblemDescription1}        
  \end{center}     
\end{figure}

We provided an example problem instance we created with this version of the problem in Figure \ref{Figure::ProblemDescription1}. In this problem instance, there are 3 locations, 4 UAVs, and a depot. $Tw$ here represents the time window. For example, for location-2 we have a time window [1-5] which means that an agent cannot be somewhere else between time step 1 and 5. Because of that, we can not assign UAVs 1, 2 and 3 to location-1, as it's time window collides with the time window of location-1.
\section{Method}
Well-known methods that solve VRPTW variations usually employ linear programming or genetic algorithm methods. In this section, we present a novel approach, Simplex VRP Algorithm (SVA), which can solve m-VRPTW problem instances efficiently. m-VRPTW has a combinatorial nature~\cite{journals/networks/LenstraK81} and its running time is exponential to the number of agents and number of targets. The motivation of SVA is to divide the m-VRPTW problem into several sub-problems and solve the smaller problems instead of the big problem. This reduces the number of agents and targets used for each sub-problem. By this way, we can reduce the exponential effect of the problem size to the running time. The main idea of SVA is the cluster vertices according to their time windows. After the clusters formed, it creates a new sub-problem for each cluster. Then, it converts each of these sub-problems into a transportation problem. SVA solves these transportation problems incrementally and combines them to construct the overall solution.

To measure SVA's performance, we implemented four more methods. These are brute force algorithm (BFA), greedy algorithm (GA), intlinprog heuristic (IH) and hybrid greedy algorithm (HGA). We explain the implementation details of SVA and the four alternative methods in the following subsections.
\subsection{Simplex VRP Algorithm (SVA)}
The m-VRPTW problem both have scheduling and assignment dimensions in it. The main idea of SVA is to separate scheduling and assignment parts of the problem and handle them independently. We eliminate the scheduling part of the problem by clustering targets according to their time windows. We can see these clusters as new smaller VRP problem instances. Solving the transportation problems handles the assignment dimension of the m-VRPTW problem. In the end, we solve these transportation problems incrementally and merge the solutions to constitute the solution of the VRPTW problem. The name simplex in the name of the algorithm comes from the method that we used to solve the transportation problems. Simplex algorithm is one of the most successful methods to solve linear programming problem instances. It is used to solve standard maximization problems. In this approach, we use the simplex algorithm to solve transportation problem instances. 

The SVA method does not guarantee to find optimal solutions because it does not consider all possible solutions. Notwithstanding it does not guarantee to find the optimal solution, we show that it produces good quality solutions. We present a pseudo-code for this method in Algorithm \ref{Simplexpseudocode}. We explain the clustering and the conversion methods in the following paragraphs.

\begin{algorithm}
  \caption{SVA Pseudocode}\label{Simplexpseudocode}
  \textbf{Data:} $U$, vector of UAVs\\
  \textbf{Data:} $T$, vector of targets\\
  \textbf{Data:} currentTime\\
  \textbf{Result:} Source, source vector of the transportation problem\\$//$ updated target list\\
  \textbf{Result:} Destination, vector of UAVs $//$ Updated UAV list\\
  \begin{algorithmic}[1]  
    \State $Tc \leftarrow$ Cluster ($T$)
	\While{$Tc$ NOT empty}
		\State $C \leftarrow$ Pop-one-cluster from $Tc$
		\State $Cc \leftarrow$ Convert-to-transportation($C$)
		\State solve-with-simplex-method($Cc$)
		\State update($U$,$T$,currentTime)
	\EndWhile
  \end{algorithmic}
\end{algorithm}

\begin{figure}[htpb]         
  \begin{center}    
 \includegraphics[scale=0.4]{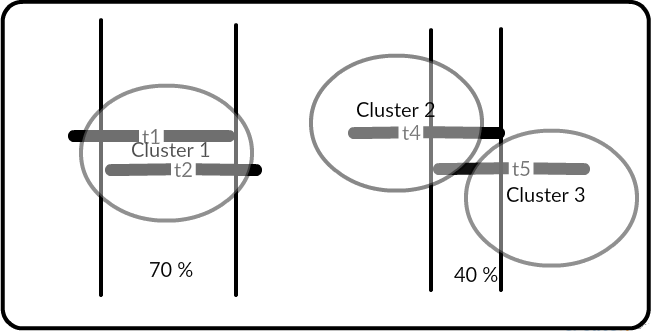}    
    \caption{UAV structures in this problem}\label{Figure::DivideAndConquer2}    
  \end{center}     
\end{figure}

The first step of clustering is to order destination locations according to their time window start times. Then, we form an initial cluster which has the first destination in it. Later, we add each destination one by one ordered by their time windows. For each destination, we compare its time window with the last formed cluster. We calculate the overlap ratio between the current destination and the last member of the cluster.
If there is 70 percent or more overlap, then we add the destination to the current cluster. Otherwise, we create a new cluster with
this destination. The process continues until no target is left. In the clusters created, each destination is a member of exactly one cluster. Figure \ref{Figure::DivideAndConquer2} helps to understand the clustering algorithm with an example. In figure \ref{Figure::DivideAndConquer2}, destinations $t1$, $t2$, $t4$, and $t5$ are ordered by their time window start times. Then, we form Cluster-1 with $t1$. Next, we add $t2$ and observe that they have a 70 percent overlap. Hence, we add $t2$ to Cluster-1. We continue adding destination locations by adding $t4$. There is no overlap between $t2$ and $t4$, so we create Cluster-2. Then, we compare $t5$ with Cluster-2. The overlap ratio between $t4$ and $t5$ is 40 percent. Hence we create another new cluster.

We run the conversion procedure for all of the clusters. In conversion procedure, first, we determine ready UAVs. We do not take UAVs currently on a mission into consideration. For each UAV, we calculate the possible reaching times to their targets. If their reaching time is smaller than or equal to the time window start of the target location, then we mark that UAV as available. Otherwise, we eliminate it. After the eliminations, next step is to cluster UAVs according to their coordinates. UAV's sharing the same destination location are counted as a cluster. UAV clusters are defined as source points for the transportation problem. Number of UAVs in each cluster shows the capacity of that source. Target locations are the destinations of the transportation problem. Needs of each target is, taken as the capacity of that destination. After this point, the problem instance becomes a transportation problem. We provide source and sink locations, their flow costs and link capacities to simplex algorithm. The solution of the transportation problem yields a minimum cost flow. For our case, results show optimal UAV-target assignments.

After SVA generates the clusters and converts each problem into a transportation problem, it solves these problems incrementally. It starts from the first cluster. After SVA finds a solution to the first cluster, it continues with the next cluster. This incremental behavior helps us to determine ready agents before solving each transportation problem. SVA solves each transportation problem by the simplex algorithm. After it solves all of the problems, it merges the solutions to generate the overall solution of the VRPTW problem.
\subsection{Brute Force Algorithm (BFA)}
We developed this algorithm to compare the results of the other implemented algorithms with the optimal solution. We use the outputs of BFA to measure the difference between the optimal result and the results of the other algorithms on small sized inputs. BFA creates all possible permutations for UAVs and targets, and then it searches these possibilities. Hence, it is guaranteed to provide optimal solutions. We can add a pre-elimination procedure to this implementation. We can use this procedure to eliminate the parts of the solution space that does not contain the optimal solution. By this way, we can increase the efficiency of BFA.
\subsection{Greedy Algorithm (GA)}
In the greedy algorithm, we sort the target locations according to their time window start times. Then, we satisfy their needs one by one. For each target location, we select a ready UAV which can reach the target with the minimum cost. Then, we send it to the target. This UAV selection process continues until we finish with the current target location. Then, we update the current time and apply the same process for the next target. 

The greedy algorithm has a faster nature than brute force algorithms. It is one of the fastest approaches to solve a problem. We are aware that greedy algorithm is not a complete approach. To have an idea about the speed performance of the generated algorithms, we implemented the greedy algorithm. So, running time of this algorithm provides a lower limit to the elapsed times of the algorithms. 
\subsection{Hybrid Greedy Algorithm (HGA)}
Another alternative algorithm we implemented is the hybrid greedy algorithm. In this one, we used an infrastructure that is similar to SVA. We used destination clustering and conversion to transportation problem ideas from SVA approach. The difference is, in this approach, we solved transportation problems with the brute force algorithm. The aim of implementing this algorithm is to observe the effect of clustering and conversion ideas to transportation problem on the running time and solution quality.
\subsection{Intlinprog Heuristic (IH)}
The final alternative method we implemented is intlinprog heuristic. This approach also uses a similar infrastructure to SVA. IH clusters the targets, and it eliminates the scheduling component of the problem. It does not convert the problem into a transportation problem. Instead of that, IH transforms the problem into a mathematical model. It generates linear equalities, linear inequalities, bounds for the solution variables and an objective function. Next, IH solves this model by a linear programming approach. To implement this algorithm, we used the Matlab optimization toolbox's intlinprog function. Data is prepared to be suitable for intlinprog's parameters. Then, we solve each of the created mathematical models by intlinprog algorithm.

To construct the mathematical model, we constituted linear equalities. To do this, we extracted the constraints of the problem. The first constraint of the problem is, only some of the UAVs are available for each sub-problem. For each of the sources, there is a fixed number of UAVs that the problem can send to a destination. The second constraint is, each target needs a fixed number of UAVs. According to these constraints, we constructed the equations. We assumed that the number of sources available is $n$ and the number of destinations available is $m$. For each source and destination, there is exactly one constraint. So, there can be $n + m$ equations for this problem. For each of the sources, we show the number of UAVs it have as: $S_{1},S_{2},S_{3} \cdots S_{n}$. This information shows how many UAVs that each of the sources can send. For each destination, the number of the UAVs it can receive is shown as: $D_{1},D_{2},D_{3} \cdots D_{m}$. With the help of this information, we created the following equation for each source $i$ to destination $j$:

\begin{equation}  \label{eq:1}
  \sum_{j=1}^{m} u_{ij} = S_{i}
\end{equation}

In equation \ref{eq:1}, $u_{ij}$ represents the number of UAVs that will be sent from source $S_{i}$ to destination $D_{j}$. For this problem no inequalities exist. Furthermore, there are no upper bounds exist for the number of UAVs it sends. Hence, we set the upper bound to infinite. It is not possible to send a negative number of UAVs to a target. So, we determined the lower bound of the solution variables as 0. The objective function of the problem defines the value that we want to minimize. In this problem, it is the total cost. The total cost is the accumulated sum of the fuels consumed by all of the UAVs. We represent the total cost by $C$. Equation 2 shows the way we calculated $C$:

\begin{equation}  \label{eq:2}
C = \Sigma ^{m} _{i=1} ( \Sigma ^{k} _{j=1} c_{ij} ) \\
\end{equation}

$c_{ij}$ is the cost of going from the current location of the UAV $j$ to the target $i$'s location. The $m$ in the formula represents the number of targets. Similarly, the $k$ in the formula represents the number of UAVs at each target.
\section{Evaluation And Results}
We compared the performance of SVA with four alternative approaches. To compare the solution qualities with the optimal solution, we implemented a brute force algorithm (BFA). We calculate solution qualities by calculating the total fuel consumptions of the provided solutions. To compare running-times with the fastest possible option, we implemented a greedy algorithm (GA). The other two algorithms are implemented to observe the impact of applied solution methods after the separation of the problem into smaller instances. We examined the performances of the algorithms with mainly three set of test beds. The first set of experiments, compare other algorithms solution qualities with BFA. Since BFA is too slow for large size inputs, this experiment only includes small-sized inputs. Then, we analyze the performance of SVA, GA, HGA, and IH with large-sized inputs. We formed the last set of experiments to see the effect of intersection ratios of targets on generated solutions. We provide the test environment, details of the data sets and the test results in the following subsections.
\subsection{Test Environment}
We developed the project on a PC which has a 64 bit 3.40 GHz Intel i7 processor. We used Ubuntu 14.04.2 LTS operating system and used Matlab programming language for implementation.


\subsection{Data Sets}
We performed the experiments with hand-crafted data and randomly generated data. Hand-crafted data contains the extreme cases. To produce all possible scenarios we used randomly generated data. For both types of tests, we performed the experiments in 3, 5, 7,10 and 25 target graphs. For the 3 and 5 target graphs, we solved the problem with three UAVs. For the 7, 10 and 25 target graphs we used five UAVs. We created five different scenarios for hand-crafted experiments. All UAVs start from a single depot. In each case, we kept target coordinates fixed and changed time windows of the targets. In randomly generated inputs, for each of the different size graphs, we produced 100 different problem instances. We reported the average of all these 100 test cases. Random graph generator takes six parameters. These parameters are the number of targets, number of UAVs, time window widths, intersection ratios of the target time windows, largest coordinate (determines environment size), and UAV needs for the targets. Random graph generator creates targets with randomly chosen coordinates. Each coordinate is between zero and largest coordinate defined. We determined the cost of travel from one target to another by the Manhattan distances between them. We created agents identically and placed them to the same depot location at the beginning. We also categorized the inputs by their time window intersection ratios. For each target, we determined the same number of needs in this study. More than one agents can share the same location (target point).

\subsection{Algorithms Used in the Tests}
In this part, we presented the algorithms we developed to solve UAV mission planning. From this point we referred to the algorithms with the abbreviations we provided in the Table \ref{Figuere::Algorithms}.

\begin{table}[htpb]
\centering
\caption{Table showing compared algorithms}
\label{Figuere::Algorithms}
\begin{tabular}{|l|l|}
\hline
{\bf Algorithm Name}    & {\bf Abbrevation}  \\ \hline
Brute Force Algorithm   & BFA   \\ \hline
Greedy Algorithm        & GA     \\ \hline
Intlinprog Heuristic    & IH     \\ \hline
Hybrid Greedy Algorithm & HGA     \\ \hline
Simplex VRP Algorithm       & SVA       \\ \hline
\end{tabular}
\end{table}

\subsection{Test Results}
\subsubsection{Experiments with brute force algorithm}
We made comparisons with BFA separately using three and five target hand-crafted graphs. We compared fuel consumption rates of the other four algorithms with BFA to see how far they are from the optimal solution. Table-\ref{Table::3hand} provides the results for three target graph. BFA took more time than others. Although the other four algorithms require acceptable times, HGA seems to be slower than the others. For the three target experiments, GA did not provide a complete scheduling for three of the test cases. Other algorithms successfully found solutions for all of the cases. To analyze fuel consumptions, we first need to explain the scenarios in the data set. For three targets case, there are only limited scenarios. In the first scenario, we created targets whose time windows totally intersected. For two inputs, target time-windows have a 100 percent intersection rate. For these two inputs, SVA and IH found the optimal solution. For the remaining cases, we separated all targets from each other. For these cases, SVA and IH acted like greedy algorithms as we expected. HGA found good solutions for these small sized inputs as expected. Because it covers most of the solution space. Another key point to realize is, SVA and IH provided same results every time. The reason of this is, these two algorithms cover the same amount of the state space. The only difference between them is their solution methods for the transportation problems. Elapsed times of these two algorithms are also similar. Elapsed time differences between GA, SVA and IH will be clearer with the larger inputs.

\begin{table}[htpb]
\centering
\caption{All algorithms are compared on 3 target hand crafted inputs}\label{Table::3hand}
\scalebox{0.6}{
\begin{tabular}{|l|l|l|l|l|l|l|l|l|l|l|}
\hline
{\bf } & \multicolumn{2}{l|}{{\bf BFA}} & \multicolumn{2}{l|}{{\bf GA}} & \multicolumn{2}{l|}{{\bf IH}} & \multicolumn{2}{l|}{{\bf HGA}} & \multicolumn{2}{l|}{{\bf SVA}} \\ \hline
Cases  & Time          & Fuel           & Time & Fuel                   & Time          & Fuel          & Time          & Fuel           & Time          & Fuel          \\ \hline
3\_1   & 3,53          & 14,12          & 0,00 & \multicolumn{1}{c|}{-} & 0,00          & 16,55         & 1,00          & 16,55          & 0,00          & 16,55         \\ \hline
3\_2   & 3,36          & 17,18          & 0,00 & 17,18                  & 0,00          & 17,18         & 1,00          & 17,18          & 0,00          & 17,18         \\ \hline
3\_3   & 3,31          & 15,84          & 0,00 & 22,78                  & 0,00          & 22,78         & 1,00          & 15,84          & 0,00          & 15,84         \\ \hline
3\_4   & 3,46          & 13,28          & 0,00 & \multicolumn{1}{c|}{-} & 0,00          & 27,06         & 1,01          & 13,28          & 0,00          & 13,28         \\ \hline
3\_5   & 3,44          & 16,11          & 0,00 & \multicolumn{1}{c|}{-} & 0,00          & 16,11         & 1,00          & 16,11          & 0,00          & 16,11         \\ \hline
\end{tabular}}
\end{table}

Table-\ref{Table::5hand} provides the results for five target graph. In these test cases, time consumption difference becomes more recognizable. BFA terminated in nearly five minutes. Where the GA, SVA, and IH worked in 0,001-0,002 seconds interval. HGA worked clearly slower. Solution qualities are similar to the three target experiment results. SVA and IH provided very similar results. When the number of time window overlaps are higher, they provided closer results to BFA compared to GA. When time windows of the targets are separate, GA, SVA, and IH provided similar behaviors. HGA's solution quality is closer to BFA than the other algorithms. For one case, HGA failed to produce a complete solution.

\begin{table}[htpb]
\centering
\caption{All algorithms are compared on 5 target hand crafted inputs}\label{Table::5hand}
\scalebox{0.6}{
\begin{tabular}{|l|l|l|l|l|l|l|l|l|l|l|}
\hline
{\bf } & \multicolumn{2}{l|}{{\bf BFA}} & \multicolumn{2}{l|}{{\bf GA}} & \multicolumn{2}{l|}{{\bf IH}} & \multicolumn{2}{l|}{{\bf HGA}} & \multicolumn{2}{l|}{{\bf SVA}} \\ \hline
Cases  & Time           & Fuel          & Time          & Fuel          & Time          & Fuel          & Time  & Fuel                   & Time          & Fuel          \\ \hline
5\_1   & 288,55         & 16,07         & 0,00          & 30,06         & 0,00          & 24,63         & 5,38  & \multicolumn{1}{c|}{-} & 0,00          & 24,63         \\ \hline
5\_2   & 285,36         & 20,18         & 0,00          & 20,18         & 0,00          & 20,18         & 5,36  & 20,18                  & 0,00          & 20,18         \\ \hline
5\_3   & 295,50         & 18,84         & 0,00          & 25,78         & 0,00          & 25,78         & 5,45  & 18,84                  & 0,00          & 25,78         \\ \hline
5\_4   & 290,00         & 18,88         & 0,00          & 30,06         & 0,00          & 31,87         & 5,40  & 20,52                  & 0,00          & 31,87         \\ \hline
5\_5   & 287,46         & 18,84         & 0,00          & 30,02         & 0,00          & 18,84         & 5,39  & 18,84                  & 0,00          & 18,84         \\ \hline
\end{tabular}}
\end{table}

\subsubsection{Experiments with large-size inputs}
\begin{figure}[htpb]       
  \begin{center}    
 \includegraphics[scale=0.7]{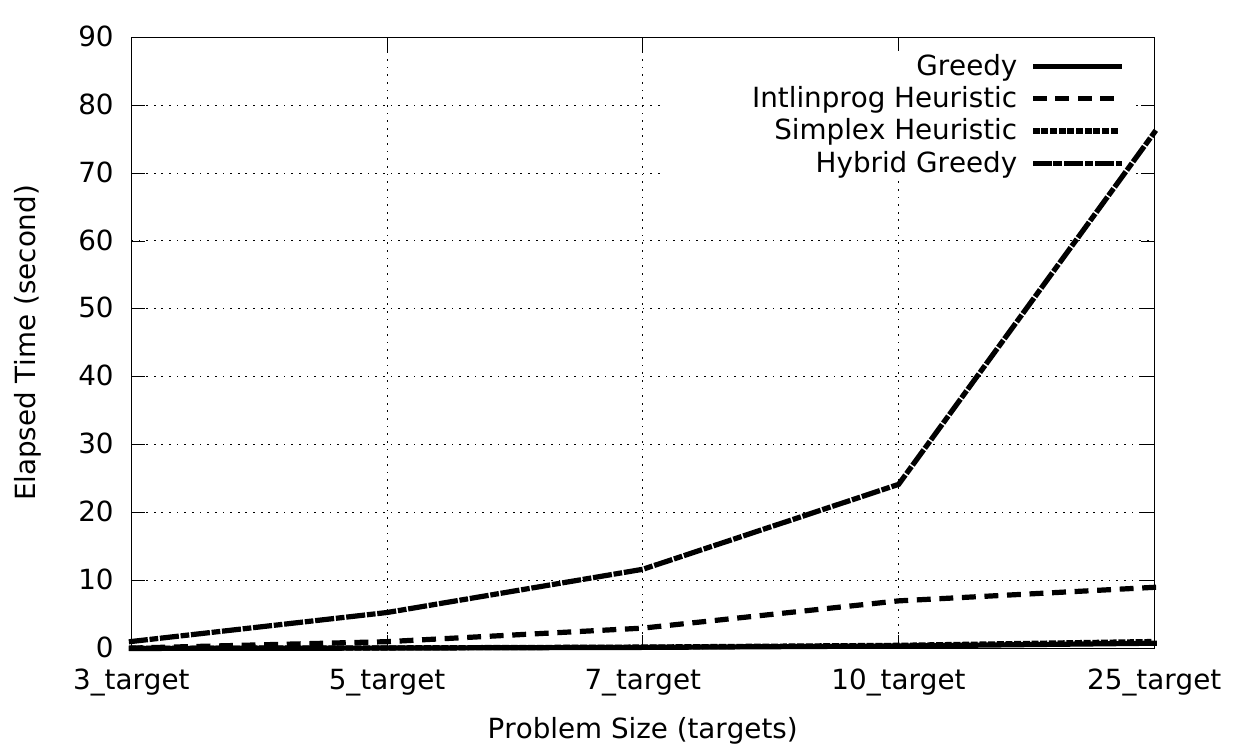}    
    \caption{Time consumptions of the algorithms for different sized randomly created inputs}\label{Figure::random}   
  \end{center}     
\end{figure}

In figure-\ref{Figure::random}, we provided the average elapsed times for each of the algorithms for different sized inputs. HGA algorithm run slower than the others. The running time was too slow for the real-time mission planning. On the other hand, other three algorithms provided adequately fast solutions for real-time planning. SVA and IH always produced the same solutions, but SVA worked faster than IH. The running time of SVA was similar compared to GA. GA worked slightly faster. The time difference is not noticeable for these sizes of inputs.

\begin{figure}[htpb]        
  \begin{center}    
 \includegraphics[scale=0.7]{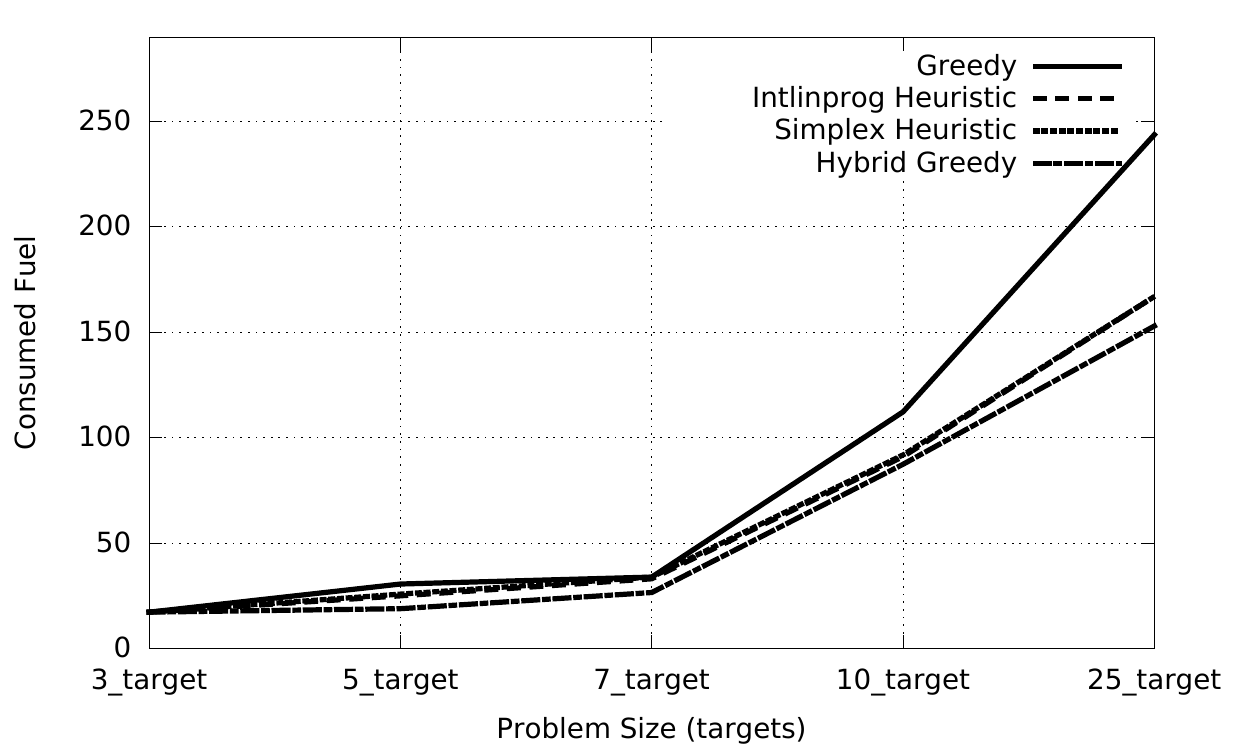}    
    \caption{Fuel consumptions of the algorithms for different sized randomly created inputs}\label{Figure::random2} 
  \end{center}     
\end{figure}

Figure-\ref{Figure::random2} presents the fuel consumption quantities of the algorithms for larger problem instances. HGA finds smaller cost results than others.  As the input size gets larger, the difference between consumed fuel amounts of GA and SVA became more apparent. Results of the SVA is better than GA's provided results for large input sizes. The consumptions of IH and SA are continues to be same, as expected. 

\begin{figure}[htpb]        
  \begin{center}    
 \includegraphics[scale=0.7]{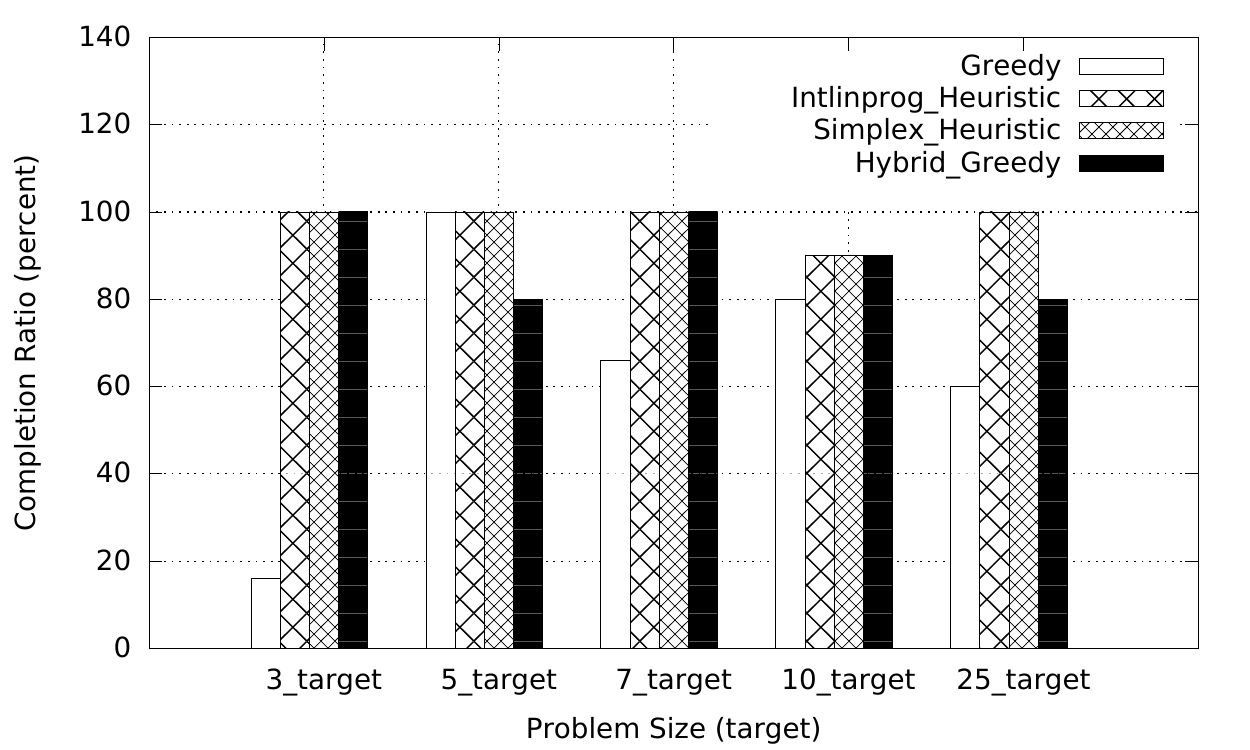}    
    \caption{Scheduling completion ratios of the algorithms for different sized randomly created inputs}\label{Figure::random3}    
  \end{center}     
\end{figure}

Figure-\ref{Figure::random3} presents the percentages of completion of the algorithms. We calculate these rates by counting the number of successful solutions that the algorithms provided. When the number of intersections between target time windows increases, greedy choices can yield us to incomplete solutions. In three target experiments, most of the cases include high overlap ratios. Hence, GA has a lower success rate compared to others. One of the critical elements of this study is to find complete solutions for large-sized problems. SVA and IH seem to satisfy this need. For the larger input sizes, we can realize that HGA and GAs overall percentages are lower than SVA and IH. SVA and IH seem to find more reliable solutions than the other two algorithms. 

\subsubsection{The effect of time-window intersection ratios on the performances of the algorithms}
We analyzed the performance of the algorithms in three different scenarios. In the first scenario, we randomly created a problem set has up to thirty percent of time-window intersections. Then we created a test set that has between thirty percent and seventy percent time-window intersections. In the last section of this experiment, we created a problem set that has between seventy percent and a hundred percent time-window intersections. For each of the sets, we used a fixed problem size having ten targets. Furthermore, for each of the sets, we employed a hundred test-cases (a mixture of randomly-created and hand-crafted inputs).

\begin{figure}[htpb]     
 \begin{center}  
 \includegraphics[scale=0.7]{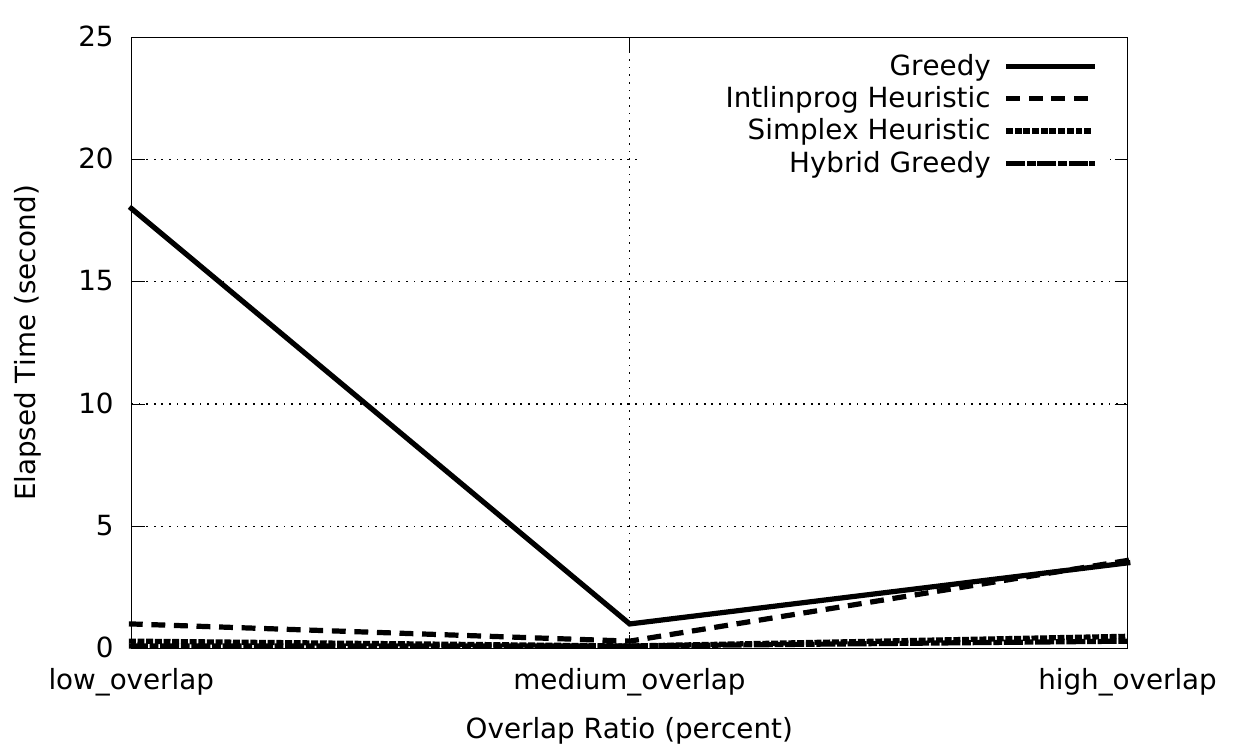}    
    \caption{Effect of time-window intersection ratios on the running time of the algorithms.}    \label{Figure::overlap1}  
 \end{center}          
\end{figure}

Figure-\ref{Figure::overlap1} provides an overview of the elapsed times of the algorithms for ten target problem size. When the problem has a low number of time window intersections, the number of decision points increases. At each decision point, algorithms have to solve a transportation problem. Using brute force search several times is costly. Therefore, HGA works much slower than the other algorithms in this case.  When there is a medium number of intersections, the condition is optimal for HGA. First of all, the number of decision points is low. Moreover, the number of UAVs and targets that will exist in the transportation problem are likely to be low. In this condition, the complexity of the transportation algorithms decreases. Hence, brute force can solve this problem faster than the low intersection case. For the high number of time window intersections, the number of decision points decreases. On the contrary, in each transportation problem, number of targets and UAVs are likely to be high which adds complexity to the transportation problems. For the other algorithms, this complexity change does not affect the running time of the algorithm. GA worked fast for all of the input types. Its performance is not affected by the intersection rate changes. IH worked fast for the low-intersection and medium-intersection problems. For the cases having high-intersection rates, the efficiency of the algorithm decreased compared to other two cases. In the high intersection cases, there is a possibility to have a low number of decision points. This situation produces, low number of large-sized transportation problems. IH  is not a very fast approach to solve the transportation problem. Hence, for the large-sized transportation problems, its running time becomes slower.   The simplex algorithm works faster than intlinprog heuristic for solving transportation problems. Therefore, complexity increase in the high intersection scenario did not affect SVA as much as IH. Still, it is possible to recognize a small increase of running time of SVA in the high intersection cases.

\begin{figure}[htpb]   
 \begin{center}  
 \includegraphics[scale=0.7]{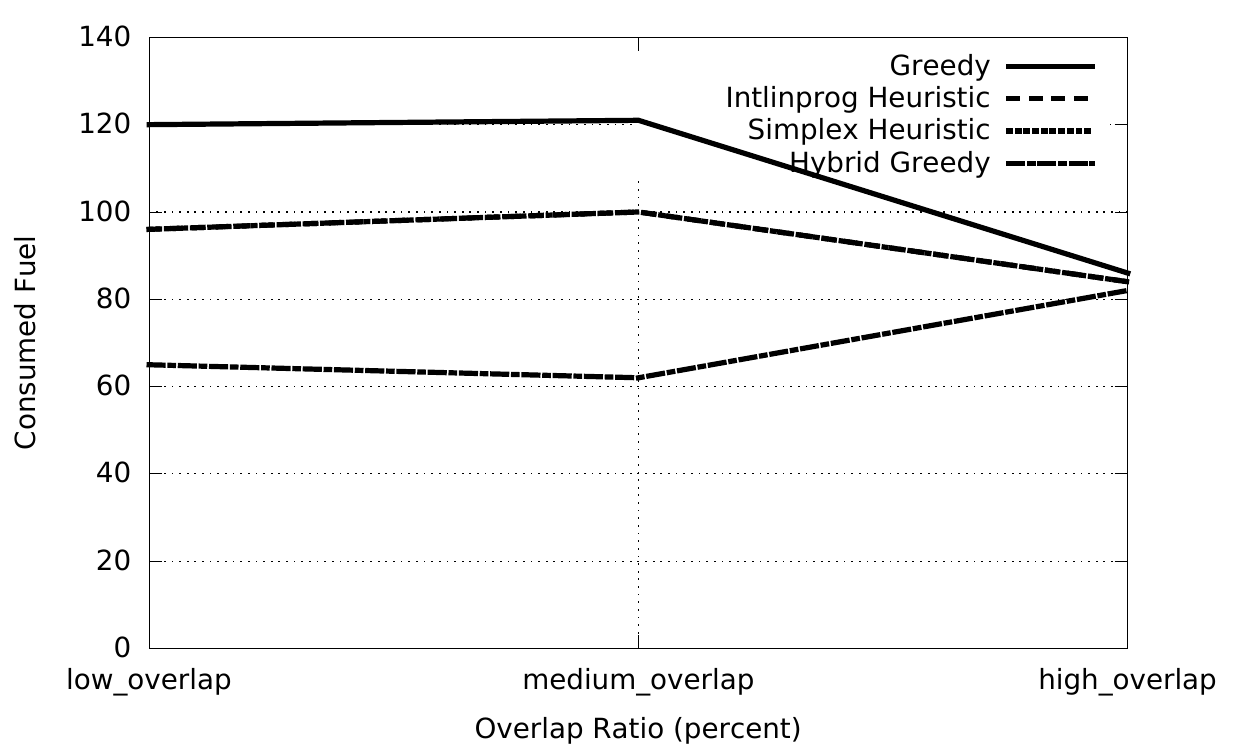}   
    \caption{Effect of time-window intersection ratios on the solution qualities of the algorithms.}\label{Figure::overlap2}  
 \end{center}          
\end{figure}

In the figure-\ref{Figure::overlap2}, we provided the fuel consumption levels of the algorithms. GA provided less optimal solutions than the others. Moreover, IH and SVA found better quality solutions than HGA. In the high intersection case, all of the algorithms performed similarly. The first reason of this is, all of the algorithms have a similar infrastructure. The second reason is, creating less number of transportation problems. For the smaller intersection cases, algorithms created much more transportation problem instances. With each transportation problem, the effect of optimality differences echoed. This situation increased their total fuel consumption differences. Since in high intersection experiments there are fewer transportation problems, this difference decreased.

\begin{figure}[htpb]   
 \begin{center}  
 \includegraphics[scale=0.7]{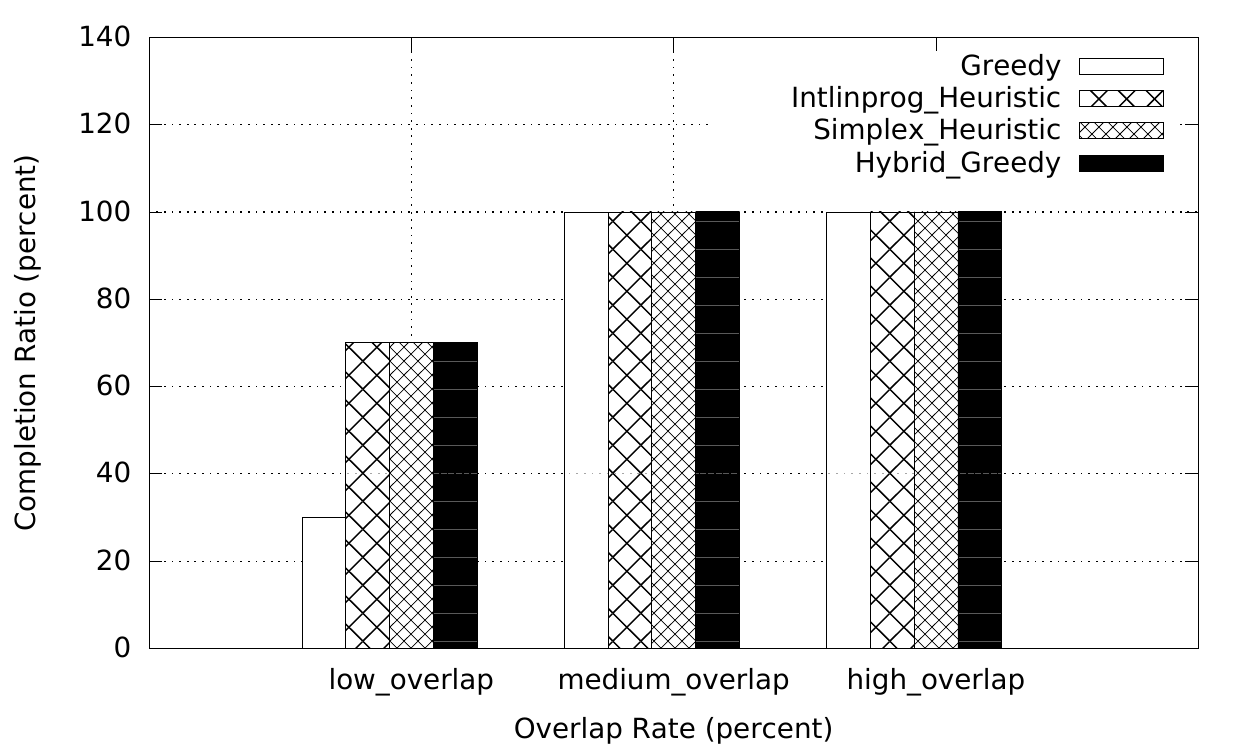}    
    \caption{Effect of time-window intersection ratios on the scheduling completion ratios of the algorithms.}\label{Figure::overlap3} 
 \end{center}          
\end{figure}

In figure-\ref{Figure::overlap3}, we provided the scheduling completion rates of the algorithms. The time window intersection rates affected GA more than the others. In the low intersection scenario, time windows of the targets are sparse. Hence algorithms generate many transportation problems. As a result, wrong choices of the GA ends up with incomplete scheduling. All of the implemented algorithms have a greedy nature. Therefore, all of them performed badly on low intersection scenarios. 

All of the provided experiments reveal that the most suitable approach amongst the provided approaches is SVA for this problem. For the small-sized inputs, HGA found close results to BFA but the running time of the algorithm is too slow to be used. Besides, HGA has a low ratio of finding a complete scheduling. GA is a faster option but its solution quality is low. IH and SVA find higher quality solutions than GA. IH and SVA always find the same results. However, IH is significantly slower than SVA.

\section{Conclusion}
We produced a system that provides security for a large area with a limited number of UAVs. We formulated this problem as a m-VRPTW instance. We introduced a fast and reliable approach to solve this problem. With the experiments, we showed that this novel approach is a better option than the provided alternatives. Our method can find good-quality solutions in a short amount of time. Furthermore, it has a high mission completion rate.

\section*{References}

\bibliography{elsevier_references}

\end{document}